\documentclass[a4paper,fleqn]{cas-dc}

\usepackage[numbers,sort,compress]{natbib}

\def\tsc#1{\csdef{#1}{\textsc{\lowercase{#1}}\xspace}}
\tsc{WGM}
\tsc{QE}

\usepackage{multirow}
\usepackage{makecell}
\usepackage{stfloats}
\usepackage{bbding}
\usepackage{amsmath}
\usepackage{algorithmic}
\usepackage{algorithm,algorithmic}
\usepackage{graphics}
\usepackage{multirow}
\usepackage[switch]{lineno}

\begin{document}
\let\WriteBookmarks\relax
\def\floatpagepagefraction{1}
\def\textpagefraction{.001}
\let\printorcid\relax

\shorttitle{DEDUCE: Multi-head attention decoupled contrastive learning to discover cancer subtypes based on multi-omics data}    

\shortauthors{L. Pan et al.}  

\title [mode = title]{DEDUCE: Multi-head attention decoupled contrastive learning to discover cancer subtypes based on multi-omics data}  

\author[a]{Liangrui Pan} \ead{panlr@hnu.edu.cn}
\author[b]{Xiang Wang} \ead{wangxiang@csu.edu.cn}
\author[c]{Qingchun Liang} \cormark[1] \ead{hm701@163.com}
\author[d]{Jiandong Shang} \ead{sjd@zzu.edu.cn}
\author[a]{Wenjuan Liu}  \ead{liuwenjuan89@hnu.edu}
\author[a]{Liwen Xu} \cormark[1]\ead{xuliwen@hnu.edu.cn}
\author[a]{Shaoliang Peng}\cormark[1]\ead{slpeng@hnu.edu.cn}

\affiliation[a]{organization={College of Computer Science and Electronic 
							  Engineering, Hunan University},
				city={Changsha},
				postcode={410083}, 
				state={Hunan},
				country={China}}

\affiliation[b]{organization={Department of Thoracic Surgery, The second xiangya hospital, Central South University},
	city={Changsha},
	postcode={410083}, 
	state={Hunan},
	country={China}}

\affiliation[c]{organization={Department of Pathology, The second xiangya hospital, Central South University},
		city={Changsha},
		postcode={410083}, 
		state={Hunan},
		country={China}}
		
\affiliation[d]{organization={National Supercomputing Center in Zhengzhou, Zhengzhou University},
	city={Zhengzhou},
	postcode={450001}, 
	state={Henan},
	country={China}}

\cortext[1]{Corresponding author}

\begin{abstract}
\textbf{Background and Objective:} Given the high heterogeneity and clinical diversity of cancer, substantial variations exist in multi-omics data and clinical features across different cancer subtypes. 

\hangafter 1
\noindent
\textbf{Methods:} We propose a model, named DEDUCE, based on a symmetric multi-head attention encoders (SMAE), for unsupervised contrastive learning to analyze multi-omics cancer data, with the aim of identifying and characterizing cancer subtypes. This model adopts a unsupervised SMAE that can deeply extract contextual features and long-range dependencies from multi-omics data, thereby mitigating the impact of noise. Importantly, DEDUCE introduces a subtype decoupled contrastive learning method based on a multi-head attention mechanism to simultaneously learn features from multi-omics data and perform clustering for identifying cancer subtypes. Subtypes are clustered by calculating the similarity between samples in both the feature space and sample space of multi-omics data. The fundamental concept involves decoupling various attributes of multi-omics data features and learning them as contrasting terms. A contrastive loss function is constructed to quantify the disparity between positive and negative examples, and the model minimizes this difference, thereby promoting the acquisition of enhanced feature representation.

\hangafter 1
\noindent
\textbf{Results:} The DEDUCE model undergoes extensive experiments on simulated multi-omics datasets, single-cell multi-omics datasets, and cancer multi-omics datasets, outperforming 10 deep learning models. The DEDUCE model outperforms state-of-the-art methods, and ablation experiments demonstrate the effectiveness of each module in the DEDUCE model. Finally, we applied the DEDUCE model to identify six cancer subtypes of AML. 

\hangafter 1
\noindent
\textbf{Conclusion:} In this paper, we proposed DEDUCE model learns features from multi-omics data through SMAE, and the subtype decoupled contrastive learning consistently optimizes the model for clustering and identifying cancer subtypes. The DEDUCE model demonstrates a significant capability in discovering new cancer subtypes. We applied the DEDUCE model to identify six subtypes of AML. Through the analysis of GO function enrichment, subtype-specific biological functions, and GSEA of AML using the DEDUCE model, the interpretability of the DEDUCE model in identifying cancer subtypes is further enhanced.

\end{abstract}



\maketitle

\section{Introduction}
Cancer is a complex disease characterized by high heterogeneity, leading to substantial individual variances in molecular features, drug response, and survival duration among patients of the same cancer type. Theoretically, single-omics studies can efficiently offer precise analysis of research subjects. Currently, single-omics has emerged as a vital research tool in life sciences, finding extensive applications in genomics, proteomics, and related fields. With advancements in omics research, multi-omics analysis technology integrates unbiased analyses of genomic, epigenomic, transcriptomic, proteomic, and metabolomic systems. This method deepens our understanding of interrelationships and regulatory mechanisms among molecules within biological organisms and phenotypes \cite{jenssen2001literature}. For instance, in COVID-19 mechanism research, the pathogenesis of COVID-19 is explored through integrating SARS-CoV-2 databases, genomic data, transcriptomic data, microbiome data, and drug treatment information of COVID-19 patients \cite{hu2021bioinformatics}.

The era of "precision medicine" has dawned with the advancement of high-throughput sequencing technology. Extensive biomedical data is expanding rapidly and being curated in public databases, including The Cancer Genome Atlas (TCGA) \cite{robert67jensen}, Gene Expression Omnibus (GEO) \cite{edgar2002gene}, cBioPortal \cite{gao2013integrative}, UCSC Xena \cite{goldman2018ucsc}, etc. As an illustration, TCGA represents a monumental initiative to amass genomic, methylation, transcriptomic, proteomic, and clinical data from thousands of patients across more than 20 types of cancer \cite{tomczak2015review}. The genome harbors genetic risk factors and disease-causing genes associated with the onset and progression of cancer. Deviations in DNA methylation can result in gene silencing or overexpression, thereby impacting the normal function and growth regulation mechanisms of cells. Transcriptomic data offer a comprehensive perspective on gene expression in cancer cells, unveiling crucial signaling pathways and targets linked to cancer development. Comprehensive data can assist researchers in comprehending the heterogeneity of recorded biological processes and phenotypes from various perspectives. Multi-omics data can help provide patients with customized treatment plans, early disease diagnosis, personalized drug selection, identification of disease subtypes, and prediction of patient prognosis, etc. However, after obtaining a large amount of data through high-throughput sequencing technology, the integration and standardization of data, along with ethical and privacy considerations, as well as cost and accessibility, pose considerable challenges in extracting valuable information from high-throughput data.

Integrating multi-omics data from cancer patients, predicting clinical phenotype characteristics (e.g., survival time, molecular subtypes, drug response), and uncovering the underlying biological mechanisms influencing clinical phenotypes pose significant challenges in clinical research. This area represents a focal point and a formidable challenge in the research landscape \cite{liu2021three,seyednasrollah2015comparison,xia2015metaboanalyst,chong2018metaboanalyst,adam2020machine,tran2021deep}. Furthermore, the analysis of cancer multi-omics data is hindered by challenges, including issues related to data quality, dataset size, data heterogeneity, and analytical methods. These challenges result in misdiagnosis, missed diagnoses, underutilization of data, repeated verification issues, and hinder the ability to administer precise treatment \cite{malone2020molecular,capobianco2022high}. Recently, various deep learning algorithms have found widespread application in cancer multi-omics data research \cite{chaudhary2018deep,picard2021integration}. For instance, initially, 16 representative deep learning methods have been employed for classifying and clustering multi-omics datasets. These methods encompass various architectures such as fully connected neural networks (FCNN), convolutional neural networks (CNN), graph neural networks (GCN), autoencoders (AE), Capsule Network (CapsNet), and Generative Adversarial Network (GAN) designed for processing cancer multi-omics data \cite{leng2022benchmark}. Secondly, an end-to-end multi-modal deep learning model (scMDC) has been developed to represent various data sources and simultaneously learn deep embedded latent features for cluster analysis \cite{lin2022clustering}. Similarly, a unified multi-omics data multi-task deep learning framework (OmiEmbeded) is designed to facilitate dimensionality reduction, multi-omics integration, tumor type classification, phenotypic feature reconstruction, and survival prediction \cite{zhang2021omiembed}. Additionally, a scalable and interpretable multi-omics deep learning framework for cancer survival analysis (DeepOmix) extracts the relationship between clinical survival time and multi-omics data to predict prognosis \cite{zhao2021deepomix}. Similarly important, a neural network method based on multi-input multi-output deep adversarial learning accurately models complex data. It employs consensus clustering and Gaussian mixture models to identify molecular subtypes of tumor samples \cite{yang2021subtype}. Finally, the Neighbourhood Component Analysis (NCA) algorithm is utilized to select pertinent features from the multi-omics dataset obtained from the TCGA and Cancer Drug Sensitivity Genomics (GDSC) databases. It is employed to develop survival and prediction models \cite{malik2021deep}. Moreover, various deep learning and machine learning methods have been applied to the diagnosis and prognosis of cancer subtypes \cite{zhao2023subtype,lipkova2022artificial}.

However, unprocessed multi-omics data is very large and contains numerous missing values and noisy data. There is heterogeneity among multi-omics data, and their measurement scales may not match, increasing the difficulty of data integration and analysis. Choosing an appropriate model to extract effective features in multi-omics data is crucial to improving the clustering performance of the model. Furthermore, most deep learning methods for clustering multi-omics data are prone to overfitting. Overfitted models tend to learn non-representative features, thereby diminishing the model's clustering performance. Moreover, the clustering outcomes of the majority of multi-omics data are currently limited to the validation stage, and there is a lack of utilization of these results for interpretable analyses of cancer subtypes.

Drawing inspiration from attention mechanisms used in natural language and computer vision for feature extraction, we employed multi-head attention mechanisms in the extraction of multiple omics data \cite{beltagy2020longformer,han2022survey,han2021transformer,pan2022noise}. This method captures long-range dependencies in each omics dataset, computing them in parallel across the entire sequence through positional encoding. This prevents the loss of information and the vanishing of gradients. We propose an unsupervised subtype decoupled contrastive learning model, named DEDUCE, based on symmetric multi-head attention encoders (SMAE) to assist in subtyping cancer multi-omics data. DEDUCE aims to leverage SMAE to extract features from multiple omics data, cluster, and identify cancer subtypes \cite{mo2022attacking}. To prevent overfitting in the DEDUCE model, we first perform data augmentation and then introduce Dropout and regularization into the SMAE. For the DEDUCE model to achieve the best clustering effect, we use a subtype decoupled contrastive learning method to jointly optimize and evaluate the similarity between samples in multiple omics datasets, facilitating subtype analysis. In the three datasets, the DEDUCE model demonstrates clear advantages over 10 deep learning models. The DEDUCE model also outperforms state-of-the-art methods in cancer multi-omics datasets. Additionally, we employed the DEDUCE model to unveil subtypes of Acute Myeloid Leukemia (AML). Through an analysis of Gene Ontology (GO) function enrichment, subtype-specific biological functions, and Gene Set Enrichment Analysis (GSEA) of AML, we further improved the interpretability of cancer subtypes within the DEDUCE model.

In summary, our innovative contributions can be outlined as follows:


\begin{itemize}
	\item [1)] The DEDUCE model adopts a unsupervised SMAE, enabling the effective extraction of long-range contextual features from multi-omics data. It facilitates the sharing of multi-omics data features, mitigating the noise interference inherent in such data, and avoids model overfitting.
	\item [2)] We propose a SMAE-based subtype decoupled contrastive learning method for clustering cancer subtypes. The DEDUCE model separates different attributes of multi-omics data features and conducts comparative learning in both feature and sample spaces simultaneously. It calculates sample similarities through joint optimization, ensuring the effectiveness of model learning.
	\item [3)] The DEDUCE model can effectively complete clustering tasks while reducing the manual workload related to feature extraction and staged clustering. We verified the effectiveness of the DEDUCE model on three multi-omics data sets and obtained the best clustering results. A large number of ablation experiments demonstrate the effectiveness of each module in the DEDUCE model. Second, the interpretability of the DEDUCE model for cancer subtype analysis is improved through bioinformatics analysis of subtypes.
\end{itemize}

\section{Methods}

\subsection{Overview}
\begin{figure*}[!t]
	\centerline{\includegraphics[width=2\columnwidth]{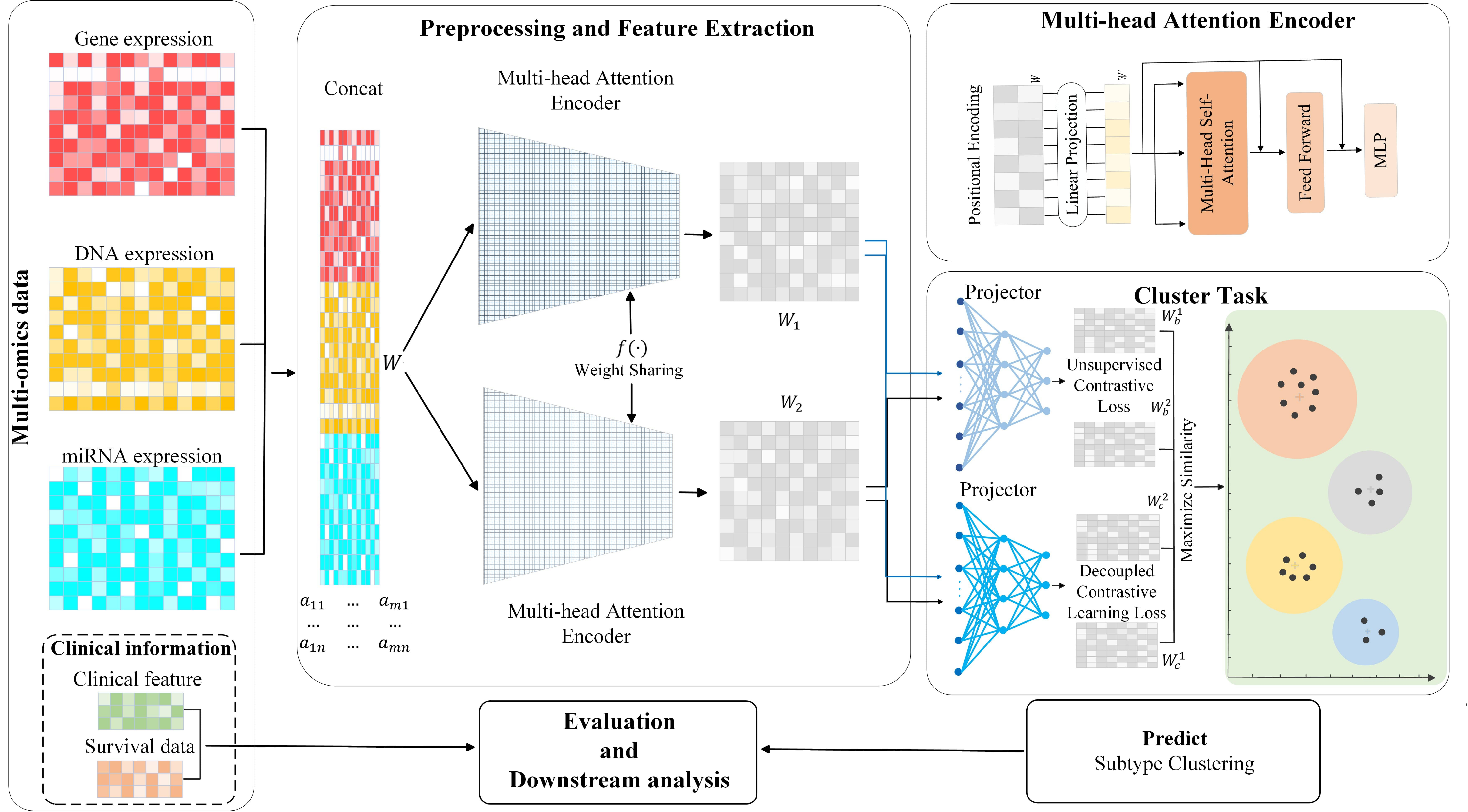}}
	\caption{Flowchart of DEDUCE model  for unsupervised subtype decoupled contrastive learning based on SMAE.}
	\label{fig:fram}
\end{figure*}

Illustrated in Figure~\ref{fig:fram}, our proposed DEDUCE model is designed to perform clustering on multi-omics data samples. Initially, we organize multi-omics data, encompassing clinical and survival-related features. Next, we preprocess and extract features from multi-omics data, involving data integration and feature extraction via a SMAE. The multi-head attention encoder conducts position encoding, linear mapping, and other operations on the multi-omics data matrix. It then transmits the feature matrix to the multi-head attention machine for feature extraction. The multi-head design of the encoder enables each head to learn different aspects of the input data, helping the model to comprehensively consider various levels of information, thereby reducing over-reliance on a specific head.  It captures various characteristics of features with different levels of importance, while suppressing attention to irrelevant features \cite{niu2021review,beltagy2019scibert}. This helps the model to utilize limited information more effectively and reduces overfitting to noisy or irrelevant information \cite{vaswani2017attention}. Next, we put dropout and regularization techniques into the model. The accuracy and robustness of the model have also been enhanced. The extracted features are transmitted to the perceptron (MLP) through the preceding layer for feature expression. Thirdly, we collectively pass the feature matrices ${W_1},{W_2}$ to the clustering tasks. During the clustering task, the feature matrices ${W_1},{W_2}$ are linearly mapped by two perceptrons. The DEDUCE model conducts clustering on multi-omics data samples employing the subtype decoupled contrastive learning method. Additionally, integrating clustering results with clinical information facilitates exploration of cancer subtyping and pathogenesis.

\subsection{Symmetric Multi-Head Attention Encoder}

Genomic data were obtained through chromosome sequencing using a third-generation sequencer. Genes are arranged in a certain order on chromosomes, and there are spatial interactions between different regions of chromosomes. Such a three-dimensional structure allows genomic data to interact with each other in different regions, forming long-range dependencies. Secondly, the long-range dependencies of transcriptome data reflect the complexity of gene expression regulation. Inspired by the great success of transformer in natural language processing, we use the symmetric multi-head attention encoders to extract features from multi-omics data in this experiment.

The multi-omics dataset has undergone preprocessing, encompassing the treatment of missing values, outliers, and duplicates. Firstly, we standardize the data to ensure uniform units across different data types, facilitating subsequent feature extraction. Secondly, we use the concat function to merge and integrate data features from diverse omics sources to enhance data coverage, augment information content, and enable comprehensive analysis. Subsequently, we shuffle the order of sample data, introduce noise to the samples, and generate training data.

In Figure~\ref{fig:fram}, the multi-omics data must be input into DEDUCE model for feature extraction and dimensionality reduction. Notably, the model does not process the multi-omics data in its sequential order.  Consequently, performing position encoding on the input multi-omics data is crucial for preserving relationships between different positions in the sequence \cite{sun2020treegen}. A fully connected layer is employed to execute the linear transformation of the input, represented as follows:

\begin{equation}
\label{eq_pe}
{y_{pe}} = {x_{pe}}\left[ {\begin{array}{*{20}{c}}
		{{W_{1,1}}}&{...{\kern 1pt} {\kern 1pt} {\kern 1pt} {\kern 1pt} {\kern 1pt} {\kern 1pt} {\kern 1pt} {W_{1,m}}}\\
		{{W_{n,1}}}&{...{\kern 1pt} {\kern 1pt} {\kern 1pt} {\kern 1pt} {\kern 1pt} {\kern 1pt} {W_{n,m}}}
\end{array}} \right] + \left[ {\begin{array}{*{20}{c}}
		{{b_{1,1}}}&{...{\kern 1pt} {\kern 1pt} {\kern 1pt} {\kern 1pt} {\kern 1pt} {\kern 1pt} {\kern 1pt} {b_{1,m}}}\\
		{{b_{n,1}}}&{...{\kern 1pt} {\kern 1pt} {\kern 1pt} {\kern 1pt} {\kern 1pt} {\kern 1pt} {\kern 1pt} {b_{n,m}}}
\end{array}} \right]
\end{equation}

Where, ${x_{pe}}$ represents the encoding vector for each position, $\left[ {\begin{array}{*{20}{c}}
		{{W_{1,1}}}&{...{\kern 1pt} {\kern 1pt} {\kern 1pt} {\kern 1pt} {\kern 1pt} {\kern 1pt} {\kern 1pt} {W_{1,m}}}\\
		{{W_{n,1}}}&{...{\kern 1pt} {\kern 1pt} {\kern 1pt} {\kern 1pt} {\kern 1pt} {\kern 1pt} {\kern 1pt} {W_{n,m}}}
\end{array}} \right]$ is the feature weight after multi-omics data integration., and $\left[ {\begin{array}{*{20}{c}}
{{b_{1,1}}}&{...{\kern 1pt} {\kern 1pt} {\kern 1pt} {\kern 1pt} {\kern 1pt} {\kern 1pt} {\kern 1pt} {b_{1,m}}}\\
{{b_{n,1}}}&{...{\kern 1pt} {\kern 1pt} {\kern 1pt} {\kern 1pt} {\kern 1pt} {\kern 1pt} {\kern 1pt} {b_{n,m}}}
\end{array}} \right]$ is the bias vector of each feature weight. The multi-head attention encoder can enhance its ability to capture relationships among different positions in sequence data through linear position encoding, consequently improving the model's performance.

Feature extraction involves employing a SMAE. The tensor matrix ${W_{mn}}$, formed by concatenating multi-omics data, serves as input for the multi-head attention mechanism, calculating features using multiple heads. Specifically, ${W_{mn}}$ is split along its last dimension into small feature vectors, each referred to as a head. Given the constraints of computing resources and task complexity, the multi-head attention mechanism employs 80 heads ($h$). Each head utilizes a dot product attention mechanism to compute attention weights relative to other heads. The output vector of self-attention for each head is represented as \cite{tay2021synthesizer}:

\begin{equation}	
{y_i} = \sum\nolimits_j {{a_{ij}}} {v_j} = \sum\nolimits_j {\frac{{{{\exp (s({k_i},{q_j})} \mathord{\left/
					{\vphantom {{\exp (s({k_i},{q_j})} {\sqrt {{d_k}} }}} \right.
					\kern-\nulldelimiterspace} {\sqrt {{d_k}} }})}}{{\sum\nolimits_n {{{\exp (s({k_i},{q_n})} \mathord{\left/
						{\vphantom {{\exp (s({k_i},{q_n})} {\sqrt {{d_k}} }}} \right.
						\kern-\nulldelimiterspace} {\sqrt {{d_k}} }})} }}} {v_j}
\end{equation}

Where ${k_i}$, ${q_j}$ and ${v_j}$ represent the key, query and value of the input feature, where $i$ represents the index of the key vector and $j$ represents the index of the query and value vector. $n$ represents the index of the query vector. $\sqrt {{d_k}} $ is the scaling factor of dimension ${d_k}$ to control the numerical range of the attention weight. ${a_{ij}}$ represents the attention weight. The attention output ${y_i}$ is obtained by multiplying the attention weight by the corresponding value ${v_j}$ and summing all $j$. The multi-head attention mechanism conducts multiple rounds of self-attention on the original input sequence. The outcomes of each attention round are concatenated and linearly transformed to yield the final output result. The computational process can be represented as follows:
\begin{equation}
\begin{array}{l}
	MultiHead(q,K,V) = Concat(hea{d_1}, \cdots ,hea{d_n}){W^O}\\
	hea{d_n} = Attention(qW_n^q,KW_n^K,VW_n^V),W_n^q \in {R^{{d_{\bmod el}} \times {d_k}}},\\
	W_n^K \in {R^{{d_{\bmod el}} \times {d_k}}},W_n^V \in {R^{{d_{\bmod el}} \times {d_v}}},W_n^O \in {R^{{d_{\bmod el}} \times h{d_v}}}
\end{array}
\end{equation}

The multi-head attention mechanism builds attention layers determined by the parameter $h$. In the forward propagation process, the feature matrix is input to the feedforward module's input layer. Each neuron in the input layer represents a feature, specifically, a column of the feature matrix. Each neuron assigns weights to its input, incorporates bias, computes output via an activation function, and transmits the output to the subsequent layer of neurons. Ultimately, the output layer generates the feature matrix.

Sharing weight matrices is a common method for achieving feature sharing in multi-modal data. Since the learned weight features in SMAE remain constant during feature extraction, weight sharing can be utilized in feature mapping. Moreover, in backpropagation, SMAE can employ identical values to update weight gradients due to weight matrix sharing.

\subsection{Subtype Decoupled Contrastive Learning}
The goal of clustering cancer subtypes is to categorize similar cancer samples into the same subtype, minimizing distinctions among different subtypes. This aims to enhance our understanding of the biological characteristics and molecular mechanisms of cancer, enabling improved diagnosis, treatment, and prognosis for patients. Unsupervised subtype decoupled contrastive learning significantly enhances similarity in matching. In cancer subtyping, where no experimentally provided labels are available, positive and negative samples consist of pseudo-labels generated through data augmentation.

The multi-omics feature matrix $W$ undergoes simple data augmentation (feature combination) and SMAE feature extraction to generate ${W_1}$ and ${W_2}$. To mitigate information loss induced by contrastive learning, the experiment normalizes the feature matrix using a three-layer perceptron and projects it into new feature spaces $W_b^1,W_b^2,W_c^1$ and $W_c^2$. $W_b^1$ and $W_c^1$ are considered positive samples for training with $n-1$ pairs, whereas $W_b^2$ and $W_c^2$ are considered negative samples for training with $n-1$ pairs. Cosine distance represents the similarity between paired samples and is defined as follows:

\begin{equation}
d(W_{{b_i}}^1,W_{{b_j}}^2) = \frac{{(W_{{b_i}}^1){{(W_{{b_j}}^2)}^T}}}{{\left\| {W_{{b_i}}^1} \right\|\left\| {W_{{b_j}}^2} \right\|}}
\end{equation}

Where $i,j \in [1,N]$. $N$ represents the number of samples. To calculate the error of each view in $W_b^1$ and $W_b^2$, we create cross-entropy loss function $L_i^K$. Therefore, the loss function between positive and negative samples is:
\begin{equation}
L_i^K =  - \log \frac{{\exp (d(W_{{b_i}}^1,W_{{b_j}}^2) \div \tau )}}{{\sum\nolimits_{j = 1}^N {\left[ {\exp (\frac{{d(W_{{b_i}}^1,W_{{b_j}}^1)}}{\tau }) + \exp (\frac{{d(W_{{b_i}}^1,W_{{b_j}}^2)}}{\tau })} \right]} }}
\end{equation}

Where $k \in [1,2]$ represents positive and negative samples. $\tau $ is the temperature parameter in the model that controls the softness of the output. Generally, the negative-positive coupling (NPC) multiplier in the cross-entropy loss (InfoNCE) can impact model training results in two ways. First, positive samples near the anchor point are deemed more crucial, being the only available positive samples. Concurrently, the gradient of negative samples gradually diminishes. Secondly, when negative samples are distant and contain less information, the model may erroneously decrease the learning rate from positive samples. Consequently, the model emphasizes negative samples more, neglecting balanced consideration of information from both positive and negative samples. This could result in errors in processing positive samples, subsequently reducing the model's accuracy. Experimental results demonstrate that subtype decoupled contrastive learning resolves the coupling phenomenon by eliminating positive pairs from the denominator, as detailed in \cite{zhang2022decoupled}:

\begin{equation}
\begin{array}{l}
	{L_{D{C_i}}} =  - \log \frac{{\exp (d(W_{{b_i}}^1,W_{{b_j}}^2) \div \tau )}}{{\sum\nolimits_{j = 1,j \ne i}^N {\left[ {\exp (\frac{{d(W_{{b_i}}^1,W_{{b_j}}^1)}}{\tau }) + \exp (\frac{{d(W_{{b_i}}^1,W_{{b_j}}^2)}}{\tau })} \right]} }}\\
	= {{ - d(W_{{b_i}}^1,W_{{b_j}}^2)} \mathord{\left/
			{\vphantom {{ - d(W_{{b_i}}^1,W_{{b_j}}^2)} \tau }} \right.
			\kern-\nulldelimiterspace} \tau } + \\
	\log \sum\nolimits_{j = 1,j \ne i}^N {\sum\nolimits_{j = 1,j \ne i}^N {\left[ {\exp (\frac{{d(W_{{b_i}}^1,W_{{b_j}}^1)}}{\tau }) + \exp (\frac{{d(W_{{b_i}}^1,W_{{b_j}}^2)}}{\tau })} \right]} } 
\end{array}
\end{equation}

The model computes the cross-entropy loss for subtype decoupled contrastive learning by evaluating $W_b^1$ and $W_b^2$ after applying all data augmentations. This allows the model to effectively identify all positive samples in the dataset. The process is outlined below:
\begin{equation}
{L_D} = \frac{1}{N}\sum\nolimits_{i = 1}^N {{L_{D{C_i}}}} 
\end{equation}

The 'labels as features' concept is predominantly employed in contrastive clustering. This method involves encoding labels as feature vectors, which are then inputted, along with the feature vectors of data points, into the clustering model for training. By incorporating labels into the feature space, the clustering problem is redefined as a contrastive learning problem. In DEDUCE model, data points within the same cluster should exhibit closer proximity in the feature space, whereas those from different clusters should be more distant. This transformation facilitates determining the cluster to which a data point belongs by assessing its similarity to other data points. The feature matrices $W_c^1$ and $W_c^2$ utilize cosine similarity to compute the error between a pair of samples, following this procedure:
\begin{equation}
d(W_{{c_i}}^1,W_{{c_i}}^2) = \frac{{(W_{{c_i}}^1){{(W_{{c_j}}^2)}^T}}}{{\left\| {W_{{c_i}}^1} \right\|\left\| {W_{{c_j}}^2} \right\|}}
\end{equation}

Where $i,j \in [1,M]$, $M$ represents the number of samples. To calculate the error of each view in $W_c^1$ and $W_c^2$, we create a clustering loss function $S_i^K$. The loss function between each pair of positive and negative samples can be represented as:
\begin{equation}
S_i^K =  - \log \frac{{\exp (d(W_{{c_i}}^1,W_{{c_j}}^2) \div \tau )}}{{\sum\nolimits_{j = 1}^N {\left[ {\exp (\frac{{d(W_{{c_i}}^1,W_{{c_j}}^1)}}{\tau }) + \exp (\frac{{d(W_{{c_i}}^1,W_{{c_j}}^2)}}{\tau })} \right]} }}
\end{equation}

By learning all positive and negative sample pairs, the total loss function can be expressed as:

\begin{equation}
\begin{array}{l}
	{L_C} = \frac{1}{{2M}}\sum\nolimits_{i = 1}^M {(S_i^K)}  + 2\sum\nolimits_{i = 1}^M {[P(W_{{c_i}}^1)\log P(W_{{c_i}}^1)]} ,\\
	P(W_{{c_i}}^1) = \sum\nolimits_{t = 1}^N {W_{{c_t}}^1 \div \left\| {{W^1}} \right\|} 
\end{array}
\end{equation}

The symbol $P(W_{{c_i}}^1)$ represents the probability distribution for subtype clustering allocation. It signifies the majority of label features obtained after each loss calculation. The features $W_b^1$ and $W_b^2$ utilize the subtype decoupled contrastive loss function to conduct clustering operations on samples, leading to clustering label output. Furthermore, these features use the clustering loss function for operations on samples, resulting in clustering feature output. Since the clustering model undergoes end-to-end training and prediction, it is crucial to optimize both the subtype decoupled contrastive loss function and the clustering loss function concurrently during the model training process. Lastly, in the clustering task, our total loss function is:
\begin{equation}
L = {L_D} + {L_C}
\end{equation}
The details of the pseudo of DEDUCE model are listed in Algorithm 1.

\begin{algorithm}[!t]
	\small
	\caption{Pseudo of DEDUCE model.}\label{alg:alg1}
	\begin{algorithmic}
		\STATE
		\STATE \textbf{Input:} Multi-omics data ${M_1},{M_2},{M_3}$.		
		\STATE \textbf{Output:} Final Cluster subtypes and features.
		\STATE \textbf{Initialization:} Randomly initialize DEDUCE model
		\STATE  Multi-omics feature fusion: $W = concat({M_1},{M_2},{M_3})$.
		\STATE \textcolor{blue}{\# Multi-head attention learning.}  
		\STATE  \textbf{while} epoch$ <  200$  \textbf{do}
		\STATE \hspace{0.5cm} epoch = epoch$+1$;
		\STATE  \hspace{0.5cm} \textbf{for} $i$ in $n$ \textbf{do}:
		\STATE \hspace{1cm} $hea{d_n} = Attention(qW_n^q,KW_n^K,VW_n^V).$
		\STATE \hspace{1cm} $MultiHead(q,K,V)$
		\STATE \hspace{1cm} Feature weight sharing: ${W_1} \leftrightarrow {W_2}$
		\STATE \textcolor{blue}{\# Unsupervised contrastive learning.} 
		\STATE \hspace{1cm} $x \to {x_i},{x_j}$
		\STATE \hspace{1cm} ${z_i},{z_j},{c_i},{c_j} = $ SMAE $({x_i},{x_j})$  
		\STATE \textcolor{blue}{\#Compute unsupervised loss:}
		\STATE \hspace{1cm}${L_D} = \frac{1}{N}\sum\nolimits_{i = 1}^N {{L_{D{C_i}}}} $ 
	
		\STATE \textcolor{blue} {\#Compute unsupervised contrastive loss:}
		\STATE \hspace{1cm}${L_C} = \frac{1}{{2M}}\sum\nolimits_{i = 1}^M {(S_i^K)}  + 2\sum\nolimits_{i = 1}^M {[P(W_{{c_i}}^1)\log P(W_{{c_i}}^1)]} $
		\STATE \textcolor{blue} {\#Total loss:} 
		\STATE \hspace{1cm}$L = {L_D} + {L_C}$
		\STATE \textcolor{blue}{\# Find clustered subtypes and corresponding features.}
		\STATE \hspace{0.5cm}\textbf{for} $i$ in (2,7) \textbf{do}
		\STATE \hspace{1cm} subtypes, features = K-means($i$)($W$)
		\STATE \hspace{0.5cm}\textbf{end for}

	\end{algorithmic}
	\label{alg1}
\end{algorithm}

\section{Experiments}
\subsection{Datasets}
\textbf{The Simulated Dataset} is created using the InterSIM CRAN package, containing complex and interconnected multi-omics data \cite{chalise2016intersim}. It comprises DNA methylation, mRNA gene expression, and protein expression data from 100 samples, with clusters configured to 5, 10, or 15. The software generates clusters for each sample under two conditions: "equal" and "heterogeneous". In the "equal" condition, all clusters have the same size, while in the "heterogeneous" condition, the cluster sizes vary randomly. This simulated dataset closely resembles a real multi-omics dataset, where the sample proportions in each cluster can be uniform or diverse. Access to all datasets is available at \footnote[1]{https://github.com/pengsl-lab/DEDUCE}.

\textbf{The Single-cell dataset} comprises 206 samples from three cancer cell lines (HTC, Hela, and K562). It includes two types of omics data: single-cell chromatin accessibility and single-cell gene expression data. The features for these omics data types are 49,073 and 207,203, respectively \cite{liu2019deconvolution,lee2020single}. Access to all datasets is available at \footnotemark[1].

\textbf{The Cancer Multi-Omics Dataset} is derived from The Cancer Genome Atlas (TCGA) and encompasses gene expression, DNA methylation, and miRNA expression data. It includes samples from breast cancer (BRCA), glioblastoma (GBM), sarcoma (SARC), lung adenocarcinoma (LUAD), and stomach cancer (STAD) from TCGA. Additional cancer types are chosen from the baseline dataset, such as colon cancer (Colon), acute myeloid leukemia (AML), kidney cancer (Kidney), melanoma, and ovarian cancer. Access to all datasets is available at \footnote[2]{http://acgt.cs.tau.ac.il/multi\_omic\_benchmark/download.html} \cite{wang2014similarity,franco2021performance}.

\subsection{Experiments details}
The DEDUCE model is developed on the pytorch1.7.1 platform using Python3.8.5. The DEDUCE models, proposed by us, are trained with two NVIDIA V100 GPUs. Hyperparameters in model training impact performance. To optimize, six main hyperparameters, including temperature of contrastive learning loss, batch size, epochs, optimizer, learning rate, and weight decay, were adjusted. The temperature of contrastive learning loss balances category distance and controls classification confidence, improving generalization ability and accuracy. Batch size influences DEDUCE model clustering performance by affecting negative sample count. Epochs limit model training time. The optimizer adjusts neural network parameters to minimize loss function, enhancing efficiency and preventing overfitting. Learning rate determines parameter update step size, affecting convergence speed and model performance. Weight decay, a regularization technique, reduces overfitting by adding a penalty term to limit model parameter size. Adjusting one hyperparameter keeps the other five unchanged. Selected hyperparameters are detailed in Table~\ref{tab:hyper}. The DEDUCE model adjusts the loss function through feature extraction and shared weight parameters. When the decoupled contrast learning loss and clustering loss simultaneously drop to a certain value and no longer fluctuate significantly, we stop training the model early. \textcolor{red}{To ensure the fairness of the comparative experiment, the hyperparameters and training methods used by the 10 common clustering methods in the experiment are consistent with the optimal parameters and training methods of DEDUCE.}

\begin{table}[H]
	\caption{Range of hyperparameters selection and optimal values} 
	\label{tab:hyper}
	\renewcommand{\arraystretch}{1}
	\centering
	\resizebox{1\linewidth}{!}{
	\begin{tabular}{c|c|c}
		\hline
		\textbf{Hyper-parameters} & \textbf{Select range} & \textbf{Optimal value}\\
		\hline
		UCL temperature           & {0, 0.5, 1}        & 0.5\\
		DEDUCE temperature       & {0, 0.5, 1}    &1\\
		batch size            & {128,256,512}  &256   \\
		epochs        & {100, 200, 500}  & 200\\
		optimizer              & {SGD, Adam, RMSProp}       & Adam\\
		learning rate  & {3e-2, 3e-3, 3e-4}        & 3e-3\\
		\hline
	\end{tabular}}
\end{table}

\subsection{Evaluation indicators}
In clustering tasks, we use C-index, Silhouette score, and Davies Bouldin Score to assess the performance of the DEDUCE model \cite{saini2019automatic,arbelaitz2013extensive}. C-index compares the dispersion of data clustering to the total dispersion of the dataset. Ideally, minimizing the C-index value for the number of clusters is optimal for partitioning the dataset \cite{hubert1976general}. Silhouette score, a distance-based clustering evaluation metric, gauges the similarity between each object in a cluster and its assigned cluster, as well as the nearest neighboring cluster. Silhouette score ranges from -1 to 1, where 1 signifies good clustering, -1 represents poor clustering, and 0 indicates equal distance between neighboring clusters. Davies Bouldin Score, a clustering evaluation metric based on cluster centers, assesses dissimilarity between clusters and similarity within clusters. Smaller Davies Bouldin Score values indicate better clustering results.

\section{Results}
\subsection{Evaluation of the DEDUCE model on simulated datasets clustering tasks}
We compared the clustering performance of the DEDUCE model with 10 commonly used methods for clustering omics data. These methods include late fusion autoencoder (lfAE), early fusion autoencoder (efAE), late fusion denoising autoencoder (lfDAE), early fusion denoising autoencoder (efDAE), late fusion autoencoder variational Autoencoder (lfVAE), early fusion variational autoencoder (efVAE), late fusion stacked variational autoencoder (lfSVAE), early fusion stacked variational autoencoder (efSVAE), the loss function is the one with the maximum mean difference Late fused variational autoencoder (lfmmdVAE) and early fused variational autoencoder efmmdVAE with maximum mean difference \textcolor{red}{ \cite{zhou2020tafa,charte2018practical,ashfahani2020devdan,yan2018abnormal,leng2022benchmark}}. Their characteristics are as follows:
(1) efAE: First concatenate multiple omics data into a feature vector, and then use AE composed of encoder and decoder to perform feature clustering. The ReLu function is used as the activation function for all layers of the encoder and the middle layer of the decoder, and tanh is used for the last layer of the decoder.
(2) lfAE: Similar to the lfAE model, the difference is that AE extracts features from each omics data separately when processing multiple omics data.
(3) lfDAE: independently process the vector features of each omics data by adding noise to the input data to construct partially damaged data, and restore it to the original input data through encoding and decoding, and then fuse and cluster later.
(4) efDAE: Process the vector features of the early fusion of multiple omics data concatenation. The subsequent steps are the same as lfDAE.
(5) lfVAE: Similar to the efAE model, multiple omics data are processed separately, and then VAE (compared with AE, the latent vector of VAE closely follows the unit Gaussian distribution) is used for feature fusion and cluster analysis.
(6) efVAE: Similar to the lfVAE model, but at the input end of the model, multi-omics data are fused early, and then VAE analysis and feature clustering are used.
(7) lfSVAE: Compared with lfVAE, this model only uses SVAE (SVAE is a stacked VAE model, in SVAE, all hidden layers follow a unit Gaussian distribution), and other parts remain unchanged.
(8) efSVAE: Each hidden layer of the encoder is fully connected to the two output layers, and the sampling steps are the same as efVAE. At evaluation time, a multiplier similar to $\beta $\-VAE is added to the loss function.
(9) lfmmdVAE: Similar to lfVAE, VAE is used to train omics data and finally classify the features integrated by multiple omics data.
(10) efmmdVAE: VAE is also used to train omics data. Except that the loss function is different from efVAE, other parts are the same.

In clustering tasks, the experiment employed a model to extract features from simulated multiple sets of omics data, producing 5-dimensional, 10-dimensional, and 15-dimensional embeddings. The dimension of the embedding was determined based on the number of clusters in the simulated omics data. Subsequently, the k-means algorithm clustered the results of dimensionality reduction for the multiple sets of omics data. Finally, the clustering results of the samples were obtained to compare the performance of eleven unsupervised methods. 

\begin{figure*}[!t]
	\centerline{\includegraphics[width=2\columnwidth]{22.pdf}}
	\caption{C-index Silhouette score, and Davies Bouldin score of eleven unsupervised methods on simulated datasets. SS and RS represent two conditions, i.e., all clusters have the same size, and clusters have variable random sizes.}
	\label{fig:moni}
\end{figure*}

In simulating dataset clustering, we initially utilized the C-index evaluation to assess the alignment between the fusion of multiple omics data and the true clustering. A lower C-index signifies smaller distances between clustered samples, indicating better clustering model performance. Examining the results in Figure~\ref{fig:moni}, most clustering methods exhibited favorable performance. However, the DEDUCE model showcased superior C-index values of 0.002, 0.022, and 0.023 under random-sized clustering and 0.005, 0.021, and 0.014 under same-sized clustering. These values surpassed other models in this evaluation index. This superiority may stem from the multi-head attention mechanism in DEDUCE, which prioritizes local data information extraction, resulting in more significant features when dealing with multiple omics data. Furthermore, we found that the clustering performance of the DEDUCE model remained good as the number of clusters increased.
\begin{figure*}[!t]
	\centerline{\includegraphics[width=1.5\columnwidth]{3.pdf}}
	\caption{C-index, silhouette score, and Davies Bouldin score of eleven unsupervised methods on single-cell multi-omics datasets. Based on the single-cell dataset, clustering analysis was performed and three internal indicators, including C-index, silhouette score, and Davies Bouldin score, were calculated. The number of clusters was set to 3, and the k-means clustering algorithm was run over 1000 times.}
	\label{fig:scell}
\end{figure*}

The Silhouette score is computed by calculating the silhouette coefficient for each sample, quantifying the extent to which a sample is correctly assigned to its cluster. From Figure~\ref{fig:moni}, it is evident that the efVAE model exhibits a higher likelihood of accurately assigning samples to their respective clusters. Conversely, the DEDUCE model ranks third, fifth, and seventh under the condition of clustering with the same size. The suboptimal clustering effect of the DEDUCE model may be attributed to the poor quality of the simulated data, including the presence of noise and outliers. Furthermore, the uneven distribution of cluster sizes in the dataset may contribute to lower Silhouette scores. It is important to note that the Silhouette score itself has limitations, particularly in accurately evaluating clustering effects for datasets with uneven density. In Figure~\ref{fig:moni}, the efVAE model demonstrates a lower Davies Bouldin score in simulated data clustering. This observation may be attributed to the VAE encoding input data into latent vectors and learning the data distribution by generating new data from these latent vectors. In contrast, the DEDUCE model ranks 3rd, 6th, and 6th in clustering with randomly sized clusters, suggesting a low Davies Bouldin score. This could be due to the insufficient significance of data features in the simulated dataset, hindering the multi-head attention mechanism's ability to extract effective features. Additionally, we observed that the number of clusters is one of the factors influencing the Davies Bouldin score. \textcolor{red}{The underlying data supporting the result analysis are provided in detail in the supplementary materials, ensuring the transparency and reproducibility of the results.}

\begin{figure*}[!t]
	\centerline{\includegraphics[width=1.53\columnwidth]{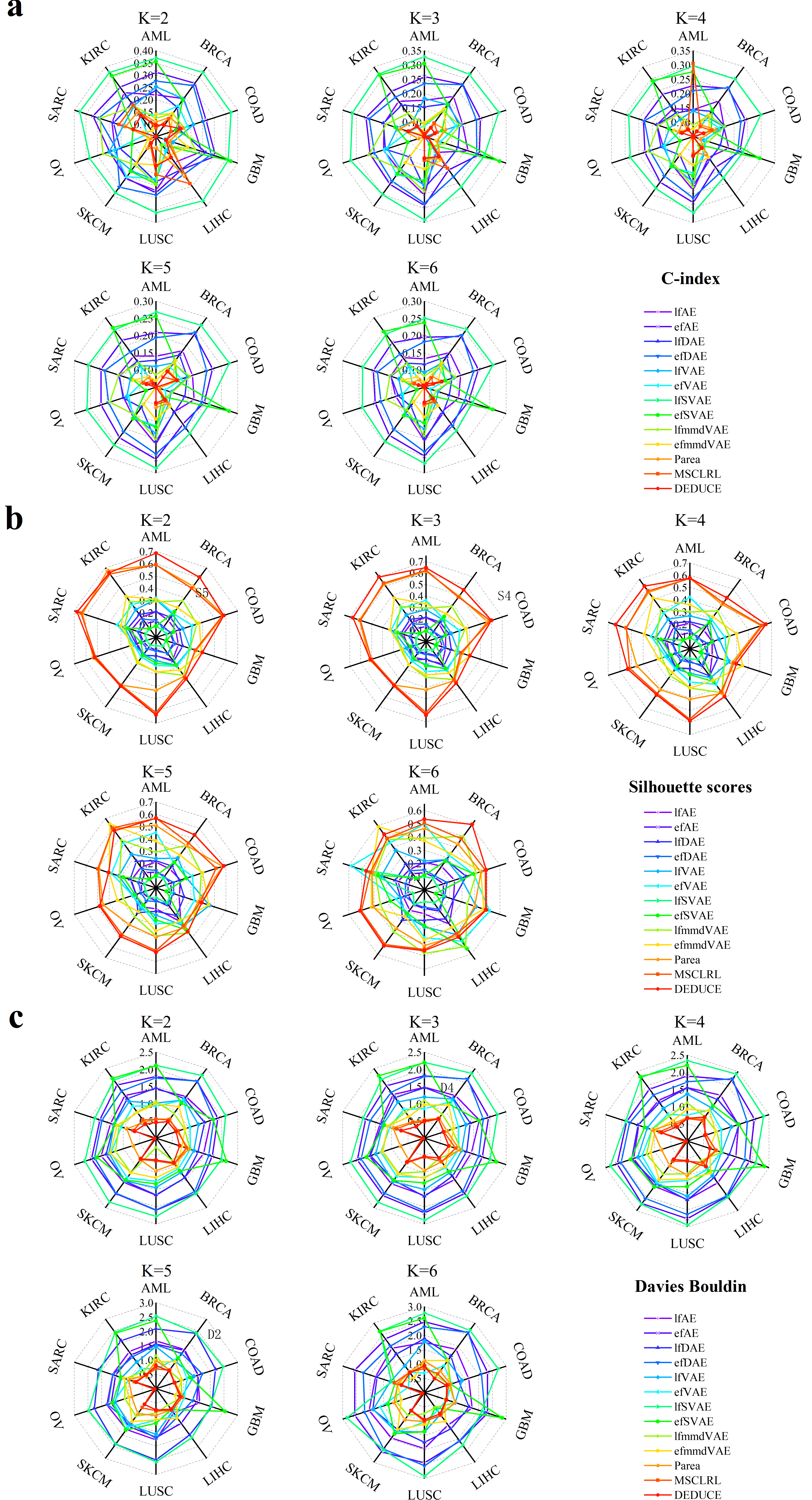}}
	\caption{C-index(a), Silhouette scores(b) and Davies Bouldin scores(c) of thirteen unsupervised methods on cancer benchmark datasets used in clustering task. Red represents the DEDUCE model.}
	\label{fig:cancer}
\end{figure*}

\subsection{Evaluation of the DEDUCE model on single-cell data clustering tasks}
In the clustering task of single-cell datasets, all models initially conduct feature fusion on multi-omics data to obtain a consolidated two-dimensional embedding. Subsequently, the k-means algorithm is employed to reduce dimensions and cluster the multi-omics data. Finally, the performance of eleven unsupervised methods is compared by consolidating clustering results into one class. The models' clustering effects are evaluated using the C-index, silhouette score, and Davies Bouldin score. As depicted in Figure~\ref{fig:scell}, the DEDUCE model attains the lowest C-index value and Davies Bouldin score, along with a higher silhouette score when clustering samples. Consequently, the DEDUCE model emerges as the optimal choice for clustering single-cell datasets. his is because single-cell data contains long sequential information, and the DEDUCE model incorporates a SMAE to focus on different local information in the data. Each head can capture varying degrees of local dependencies in the data, aiding in a better understanding of the sequential features within single-cell data. Additionally, the DEDUCE model adapts its weights by subtype decoupled contrastive learning, helping to mitigate issues such as gradient vanishing and exploding during model training when learning features from single-cell data.

\subsection{Evaluation of the DEDUCE model on cancer dataset clustering tasks}
Cancer multi-omics data demonstrate characteristics such as high dimensionality, diversity, and noise. In clustering tasks, we initially utilized eleven unsupervised models to integrate cancer multi-omics data and derive a 10-dimensional embedding. Subsequently, the k-means algorithm was employed for clustering the multi-omics data. Given the uncertainty of the optimal cluster number, experiments were conducted with clustering numbers ranging from 2 to 6. Finally, unsupervised models were employed to cluster the samples. For comparison with state-of-the-art methods, we use Parea \cite{pfeifer2023parea} and MSCLRL \cite{ge2023multi} to compare with the DEDUCE model, and show the results together in Figure~\ref{fig:cancer}. When evaluating the self-supervised clustering model, performance metrics including C-index, silhouette score, and Davies Bouldin score were employed. As depicted in Figure~\ref{fig:cancer} (a), across all clustering experiments, the C-index of the DEDUCE model primarily concentrated in the middle of the radar chart. In radar chart coordinates, proximity to the center point indicates smaller values. Thus, the DEDUCE model's C-index value reflects nearly accurate clustering of the samples. This could be attributed to the DEDUCE model's robust ability, facilitating enhanced feature capture, improved feature extraction, and data dimensionality reduction. The clustering effects of the efmmdVAE, efVAE, and lfmmdVAE models were also commendable, serving as reference models for cancer multi-omics datasets.

From Figure~\ref{fig:cancer} (b), it is evident that the Silhouette scores of the DEDUCE model are generally high in most cancer multi-omics datasets, mainly distributed in the outer circle of the radar chart. However, the DEDUCE model achieved low values in the 2-clustering tasks of SKCM and LUSC. This may be attributed to the complexity of the cancer multi-omics data structure and the large number of data points, potentially leading to underfitting in 2-clustering. In other words, the model may not effectively capture the essential features of the dataset, resulting in confusion between data points from the two clusters after segmentation.

Davies Bouldin scores are crucial evaluation indices for analyzing the clustering effect of the DEDUCE model. Hence, we also employ Davies Bouldin scores to gauge the model's performance. As illustrated in Figure~\ref{fig:cancer}(c), in 2-clustering and 3-clustering, Davies Bouldin scores attained the smallest values. However, in 4-clustering, 5-clustering, and 6-clustering, the clustering effect of the DEDUCE model on LUCS and LIHC is slightly inferior. This may be due to overfitting when dealing with multi-omics datasets with a complex structure and fewer data points in 4-clustering, 5-clustering, and 6-clustering. In these cases, dividing the dataset into three clusters may result in unnecessary subdivisions that do not well reflect the essential characteristics of the dataset, leading to a suboptimal clustering effect.

\subsection{Ablation Study}
\begin{figure*}[!t]
	\centerline{\includegraphics[width=1.45\columnwidth]{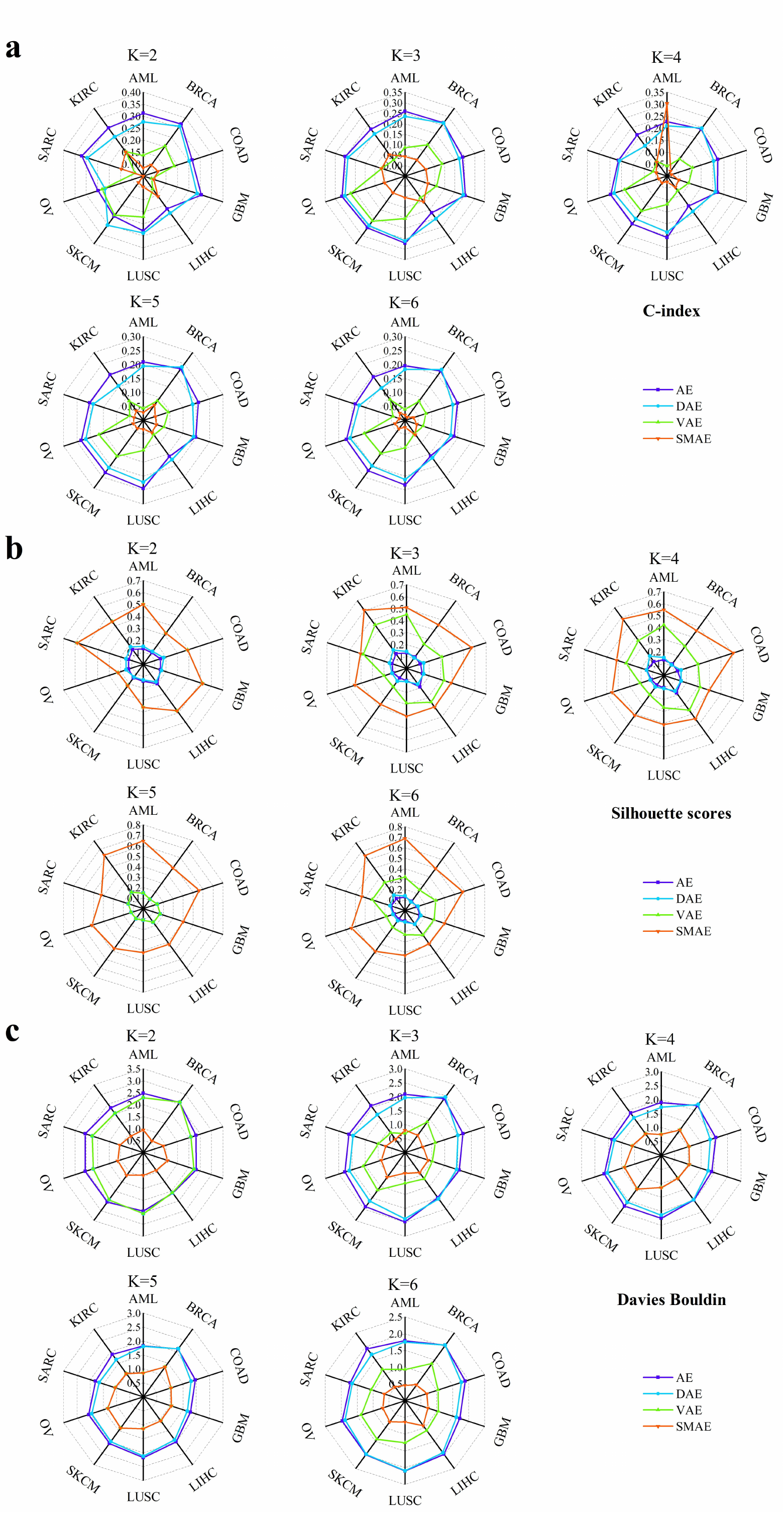}}
	\caption{Radar plot of C-index (a), Silhouette score (b), and Davies Bouldin score (c) obtained from ablation experiments in the cancer benchmark dataset. Red represents the SMAE.}
	\label{fig:smaexiaorong}
\end{figure*}

\textbf{Impact of the SMAE:} To validate the crucial role of the SMAE in the DEDUCE model, we conducted ablation experiments on the feature extraction module. The experiments involved training the DEDUCE model using AE, VAE, SVAE, and SMAE modules, and obtaining experimental results. As shown in Figure~\ref{fig:smaexiaorong}, SMAE achieved lower C-index values, higher Silhouette scores, and lower Davies Bouldin scores on most cancer datasets, indicating that SMAE can yield better results in clustering tasks. This may be attributed to SMAE's utilization of the multi-head attention mechanism, allowing it to extract both local features of cancer datasets and long-range dependencies, i.e., global features. In comparison to AE, VAE, and SVAE models, SMAE demonstrated significant advantages in feature extraction.

\begin{figure*}[!t]
	\centerline{\includegraphics[width=1.45\columnwidth]{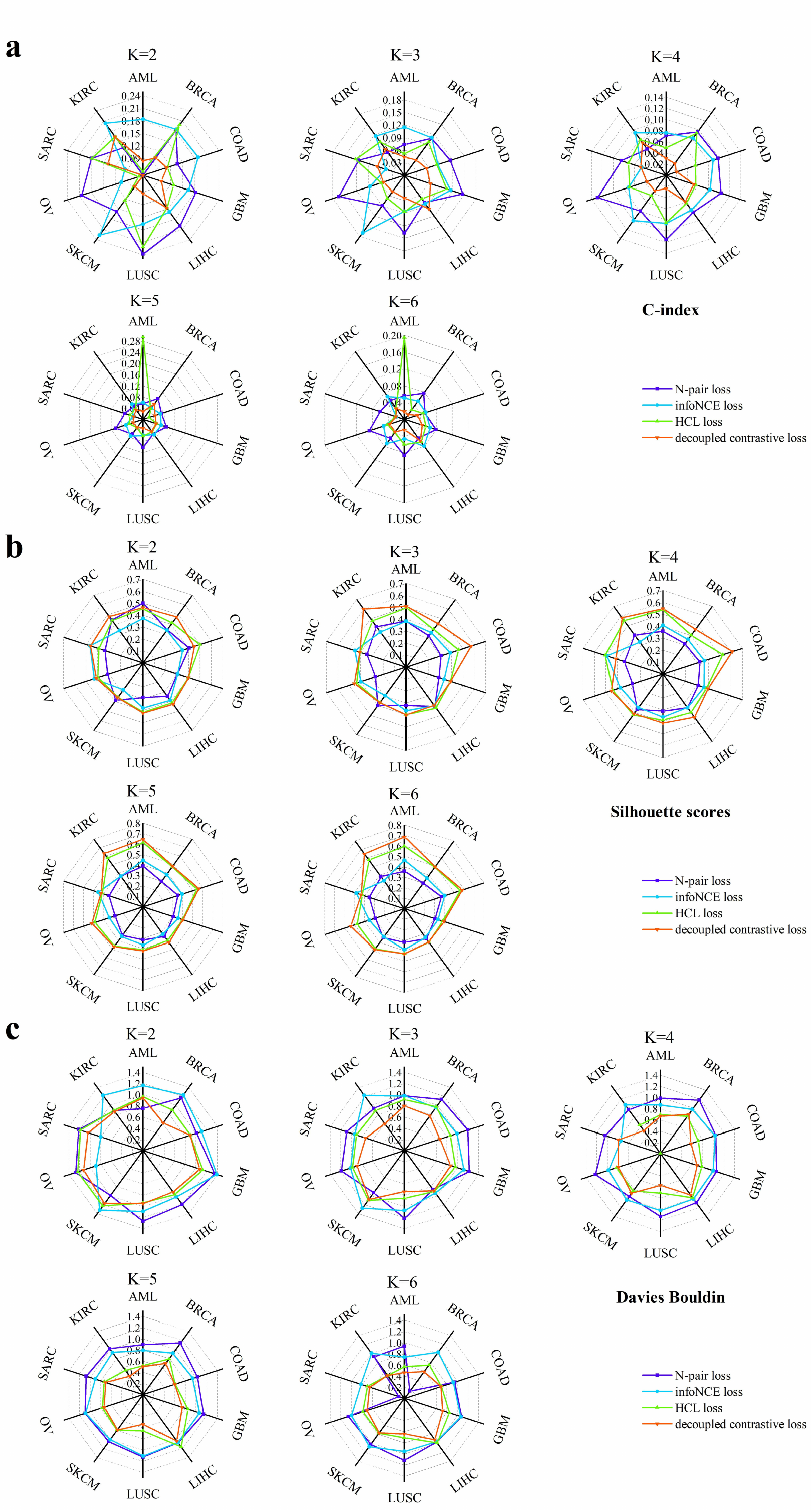}}
	\caption{Radar plot of C-index (a), Silhouette score (b), and Davies Bouldin score (c) obtained from ablation experiments in the cancer benchmark dataset. Red represents the decoupled contrastive loss.}
	\label{fig:dclxiaorong}
\end{figure*}

\textbf{Impact of the decoupled contrastive loss:} To assess the significance of decoupled contrastive learning loss in subtype decoupling contrastive learning experiments, we chose common contrastive learning losses (N-pair loss \cite{NIPS2016_6b180037}, infoNCE loss\cite{He_2020_CVPR}, and HCL loss \cite{robinson2020hard}) for comparison to demonstrate the benefits of decoupled contrastive loss. As shown in Figure~\ref{fig:dclxiaorong}, we used N-pair loss, infoNCE, HCL, and decoupled contrastive loss to optimize the DEDUCE model, respectively. In the test of multiple cancer datasets, the DEDUCE model optimized by decoupled contrastive loss achieved lower C-index values, higher Silhouette scores, and lower Davies Bouldin scores. This may be because decoupled contrastive loss improves the robustness of the model by reducing the coupling between samples, making it more generalizable. Secondly, decoupled contrastive loss can effectively address the imbalance problem of different clustering categories and increase attention to small category samples. Therefore, decoupled contrastive loss is crucial for training the DEDUCE model.

\subsection{Downstream analysis of DEDUCE model to identify cancer subtypes}
\begin{figure*}[!t]
	\centerline{\includegraphics[width=1.5\columnwidth]{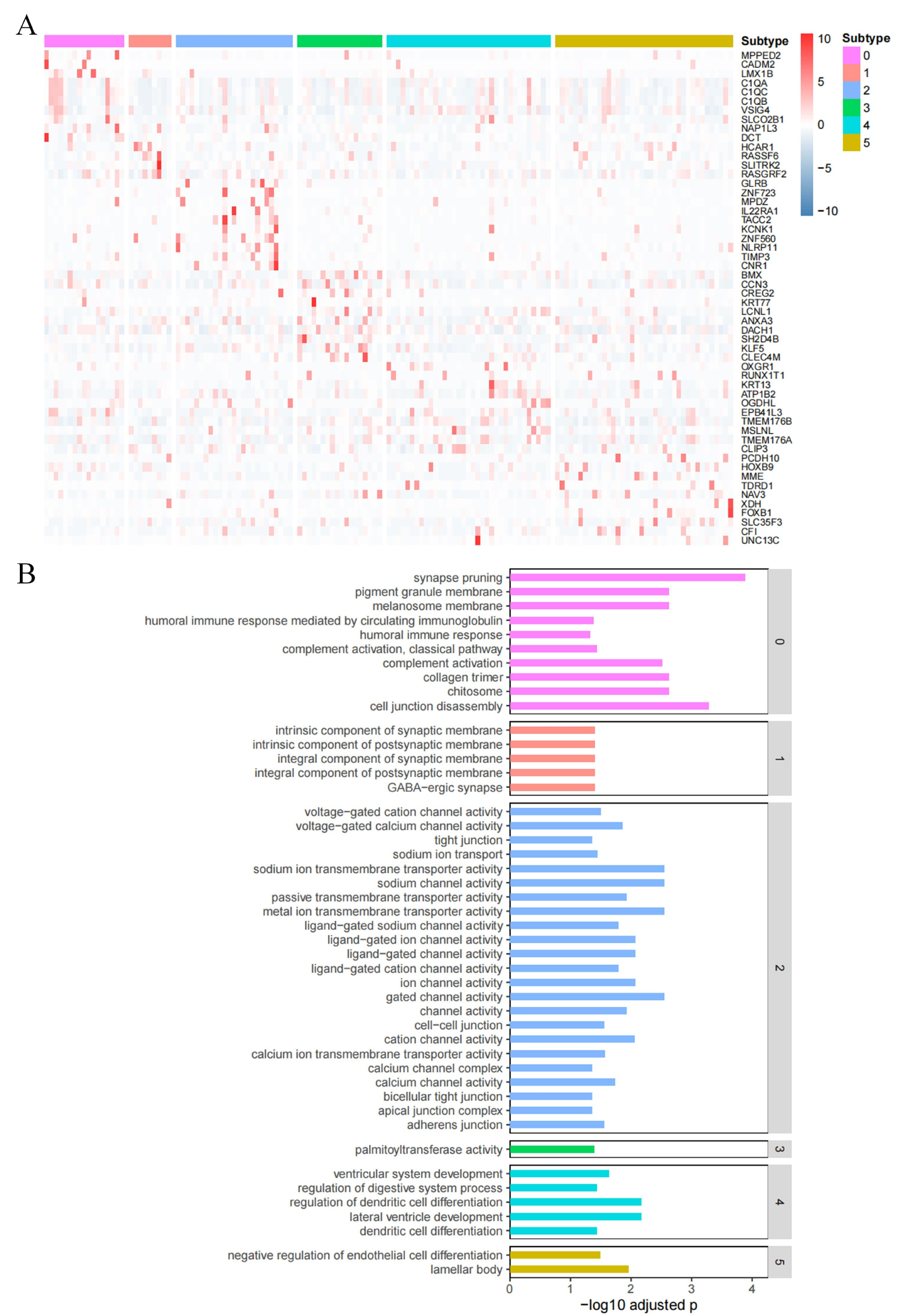}}
	\caption{Molecular characteristic differences among AML subtypes. (A) Differential expression genes of AML subtypes, heatmap color represents gene expression level (FPKM), each subtype shows the top 10 genes with differential expression fold change. (B) GO enrichment analysis of upregulated genes of each AML subtype. The y-axis represents GO-enriched terms. The x-axis represents the adjusted P-values.}
	\label{fig:tu5}
\end{figure*}

To explore the multiple cancer subtypes identified by the DEDUCE model, we chose AML, a cancer type demonstrating optimal clustering results, for investigating molecular functional distinctions. Metrics such as C-index, silhouette score, and Davies Bouldin score indicated optimal clustering efficiency at K=6. Consequently, we categorized AML samples into 6 subtypes based on the clustering results. Differential expression analysis between subtypes, using transcriptome data (comparing each subtype with all other samples), revealed specific upregulated genes for each subtype, highlighting molecular feature differences among subtypes (Figure~\ref{fig:tu5}A). For instance, complement subunit genes C1QA, C1QB, and C1QC exhibited upregulation in subtype 0, suggesting activation of the complement system in this subtype, leading to inflammation and cell lysis—factors closely associated with tumor progression.

\begin{figure*}[!t]
	\centerline{\includegraphics[width=1.5\columnwidth]{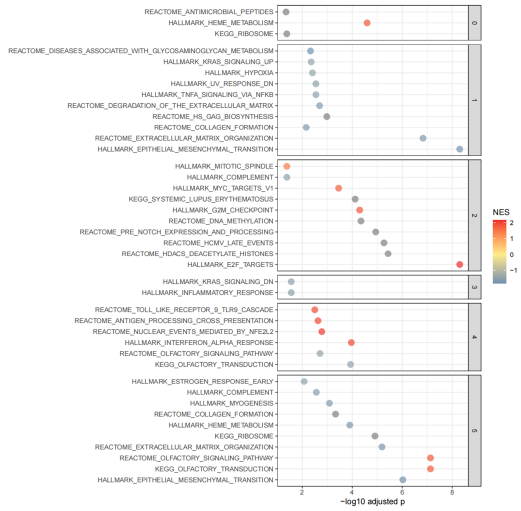}}
	\caption{GSEA results of KEGG/RECTOME pathways and cancer hallmarks in AML subtypes. The x-axis represents the adjusted P-values. The color of dots represents the enrichment score (NES).}
	\label{fig:GSEA}
\end{figure*}

\begin{figure}[!t]
	\centerline{\includegraphics[width=1\columnwidth]{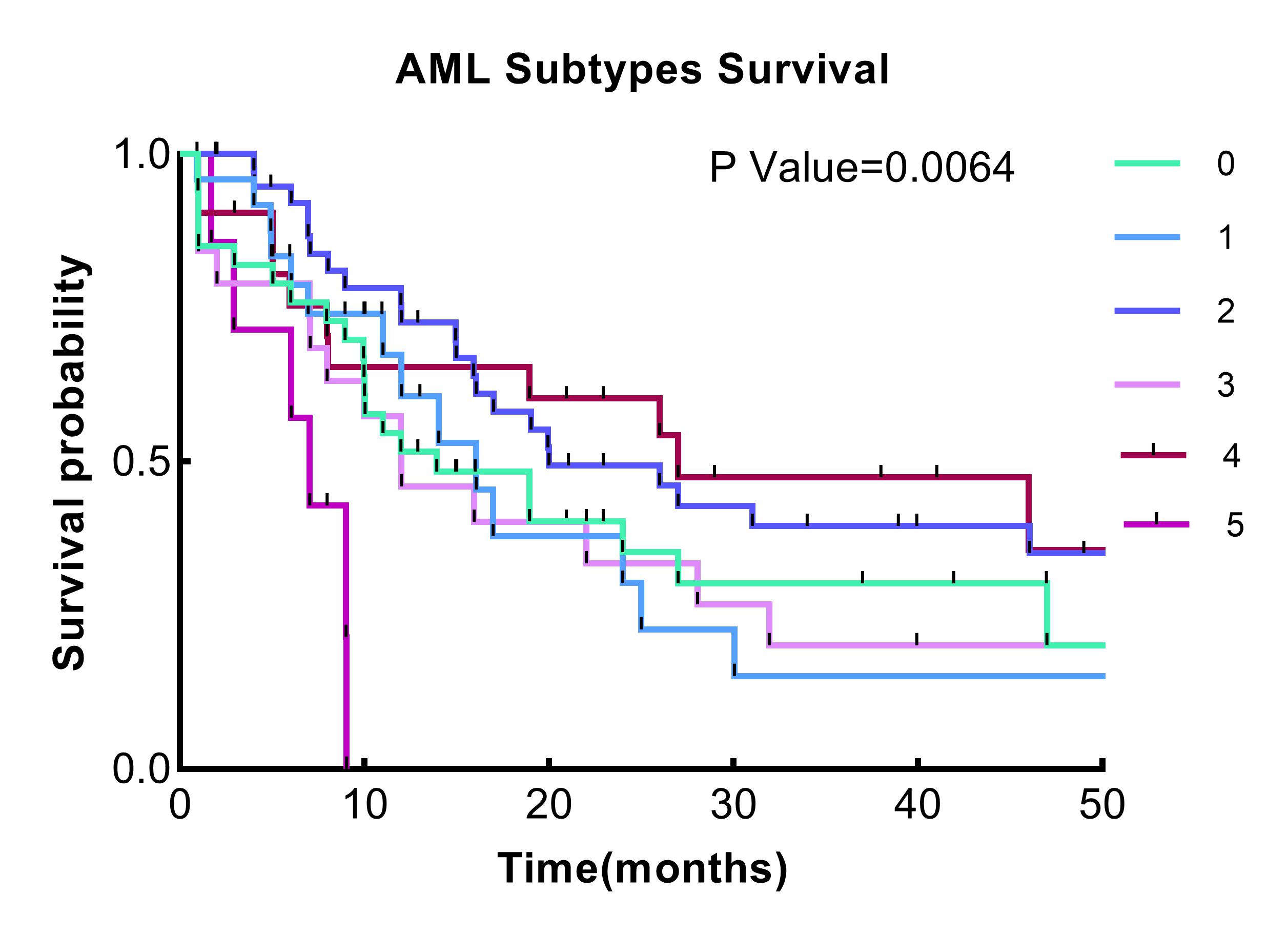}}
	\caption{Survival curves of 6 AML subtypes.}
	\label{fig:surv}
\end{figure}

Moreover, leveraging the upregulated genes of each subtype identified by the R package cluster Profiler (version: 3.14.0) for GO functional enrichment \cite{yu2012clusterprofiler}, we explored subtype-specific biological functions (Figure~\ref{fig:tu5} B). The results revealed that subtype 0's upregulated genes were primarily involved in immune response processes like humoral immune response and complement activation, along with cell junction disassembly. Differentially expressed genes in subtype 2 were mainly associated with processes such as calcium channel activity, ion channel activity, and cell-cell junction. Subtypes 4 and 5 exhibited differentially expressed genes linked to dendritic cell and endothelial cell differentiation processes, respectively. Building upon this, we employed Gene Set Enrichment Analysis (GSEA) to identify dysregulated pathways and cancer hallmarks in different subtypes (Figure~\ref{fig:GSEA}). Subtype 1 exhibited enrichment in cancer hallmarks, including EPITHELIAL MESENCHYMAL TRANSITION, HYPOXIA, and the NFKB/TNFA SIGNALING pathway. Subtype 3 showed enrichment in cancer hallmarks such as INFLAMMATORY RESPONSE and KRAS SIGNALING. Subtype 4 demonstrated enrichment in the ANTIGEN PROCESSING CROSS PRESENTATION pathway, aligning with its upregulated dendritic cell differentiation process. These biological processes illustrate the molecular functional distinctions among AML subtypes identified through the integration of multi-omics features, thereby enhancing the interpretability of cancer subtypes within DEDUCE model. such as EPITHELIAL MESENCHYMALTRANSITION, HYPOXIA, and the NFKB/TNFA SIGNALING pathway. Subtype 3 was enriched in cancer hallmarks such as INFLAMMATORY RESPONSE and KRAS SIGNALING. Subtype 4 was enriched in the ANTIGEN PROCESSING CROSS PRESENTATION pathway, corresponding to its upregulated dendritic cell differentiation process. These biological processes demonstrate the molecular functional differences of AML subtypes revealed by the integration of multi-omics features, further increasing the interpretability of cancer subtypes based on DEDUCE model. \textcolor{red}{Secondly, as shown in Figure~\ref{fig:surv}, the Kaplan-Meier survival curves and Log-rank test survival analysis for different subtypes clustered by DEDUCE revealed significant differences in survival among the six subtypes, indicating that the clustering results may have practical clinical value. These results suggest that the clustering can distinguish patient groups with different prognostic outcomes, thereby demonstrating the significance of the clustering results (P=0.0064) \cite{xie2005adjusted}.}

\section{Discussion}
\subsection{Meaning and impact}
The rapid advancement of high-throughput sequencing technology has enabled the utilization of molecular-level data for personalized medicine with unprecedented detail \cite{wang2021mogonet}. Multi-omics techniques integrate various data types, forming a more comprehensive dataset, thereby elucidating the complexity of biological systems. In this study, we propose a DEDUCE model based on the SMAE for unsupervised contrastive learning to systematically analyze three representative cancer multi-omics datasets under different contexts (simulated multi-omics dataset, single-cell multi-omics dataset, and cancer multi-omics dataset). For each dataset, we devised clustering tasks and assessed the model's performance using metrics such as C-index, Silhouette scores, and Davies Bouldin scores.

Given the superior performance of the SMAE over AE, VAE and DAE in multi-omics data feature extraction, we introduced the DEDUCE model, incorporating a SMAE, for clustering tasks. In the assessment using simulated multi-omics data, the DEDUCE model demonstrated commendable performance in C-index, but exhibited suboptimal results in Silhouette scores and Davies Bouldin scores. This discrepancy may stem from inherent issues in the dataset, such as noise, outliers, and a limited sample size. Evaluation on single-cell multi-omics data showcased the DEDUCE model achieving optimal performance indicators, attributed to its enhanced capability to capture single-cell data features, consequently enhancing clustering accuracy. In the evaluation of cancer multi-omics data, the DEDUCE model demonstrated relatively accurate clustering across most cancers. Nonetheless, an analysis of suboptimal clustering results identified potential causes, including the limited number of cancer samples leading to overfitting during model training.

The key feature of the DEDUCE model lies in its utilization of a subtype decoupled contrastive learning method, enabling the effective utilization of extensive unlabeled data for enhanced feature capture. Furthermore, subtype decoupled contrastive learning promotes feature acquisition by maximizing the similarity between positive samples, facilitating the mapping of samples from the same category into a condensed feature space and enhancing the discriminative capacity of feature representation. Extensive experimentation across three datasets validates that the subtype decoupled contrastive learning method significantly enhances the clustering performance of the model. However, it is essential to note that the implementation of subtype decoupled contrastive learning often demands large-scale sample pairing to form positive and negative sample pairs, introducing computational and memory overhead challenges, particularly in the context of multi-omics datasets. The selection of appropriate hyperparameters (such as sample selection strategy, positive and negative sample ratio) in subtype decoupled contrastive learning may necessitate specific expertise and debugging efforts to achieve optimal performance.

Leveraging the outstanding clustering capabilities of the DEDUCE model, we can employ the clustering results in conjunction with multiple omics data to scrutinize AML's GO functional enrichment, subtype-specific biological functions, and GSEA. This enhances the elucidation of the interpretability of the DEDUCE model in exploring cancer subtypes.

\subsection{Limitation}
The TCGA database used in this experiment is a multi-modal or multi-omics database of American cancer patients collected by the National Cancer Institute (NCI) and National Human Genome Research Institute (NHGRI). In the database, the number of patient samples for each cancer is relatively small, \textcolor{red}{the patient deviation is large,} and most individual cancer samples are within 1,000. \textcolor{red}{The amount of external validation data is relatively small.} Secondly, the data in this database have been collected since 2006. The relevant sequencing data were obtained using first-generation sequencers or second-generation sequencers, and the overall sequencing data's quality is not high. In summary, the quantity and quality of samples may impact the experiment results. In the future, obtaining more high-quality multi-omics data on cancer patients would benefit the later work of this study.

However, the model may overfit the noise and specific samples in the training set and not generalize well to new data. Therefore, we use data enhancement methods (shuffle sample, add noise) in the data preprocessing part to improve the model's robustness. When processing multi-omics data, the model's complexity should be reduced to avoid it trying to adapt to each data point. Overly complex models may capture subtle features in multi-omics data that are not representative of unseen data. Therefore, we only use a SMAE, Dropout, and regularization mechanism to help DEDUCE avoid overfitting. Finally, the early stopping function needs to be used during the model's training process (stop training when the loss does not change within a certain range) to alleviate the low sensitivity to new data caused by over-training. If there are more multi-omics data in the future, overfitting will be fundamentally avoided. \textcolor{red}{Moreover, the analysis of the interpretability of the model is very important for users. However, DEDUCE is based on a multi-head attention mechanism framework, and its interpretability effect is limited. It is an important research direction in the future.}

Of course, our subtypes cluster data on patients with known cancer subtypes. However, the classification criteria of some cancer subtypes have been continuously optimized in recent years, so the traditional subtype classification task cannot meet the identification or identification of new subtypes. This experiment uses the characteristics of subtype multi-omics data to cluster subtypes, and then more clustering results can be found. Through bioinformatics analysis techniques, including differential gene expression, GO enrichment analysis, and GSEA analysis, it can be determined whether it is a new subtype. It is worth noting that the maximum clustering result in this experiment was set to 6, meeting the classification criteria of most cancer subtypes. At a later stage, the results of more clusters can also be explored to discover new subtypes beyond the established cancer subtypes.

\section{Conclusion}
In this study, we propose a powerful unsupervised contrastive learning model that utilizes an attention mechanism to analyze cancer multi-omics data for identifying and characterizing cancer subtypes. Our experiments demonstrate that the SMAE effectively captures long-range relationships within each omics dataset, enabling parallel calculation of the entire sequence with position encoding. This method avoids information loss and gradient disappearance. Notably, the proposed DEDUCE model learns features of multi-omics data through the attention mechanism, and the subtype decoupled contrastive learning continuously optimizes the model for clustering and identifying cancer subtypes. This unsupervised contrastive learning method jointly optimizes the model and clusters subtypes by assessing the similarity between multi-omics data samples in both feature and sample space. The DEDUCE model exhibits a significant advantage in discovering cancer subtypes. Overall, the DEDUCE model outperform 10 deep learning models in clustering tasks. The DEDUCE model outperforms state-of-the-art methods on cancer datasets. Through ablation experimental analysis, SMAE and subtype decoupled contrastive learning can effectively improve the cancer subtype classification performance of the DEDUCE model. Lastly, we employ the DEDUCE model  to unveil six subtypes of AML. The interpretability of DEDUCE model, based on the DEDUCE model in identifying cancer subtypes, is further enhanced through the analysis of GO functional enrichment, subtype-specific biological functions, and GSEA in AML.

\section*{CRediT authorship contribution statement}
Liangrui Pan: Conceptualization, Methodology, Writing – original draft, Writing – review \& editing. Xiang Wang and Qingchun Liang: Conceptualization, Validation. Wenjuan Liu and Jiandong Shang: Visualization. Qingchun Liang, Liwen Xu and Shaoliang Peng: Supervision.

\section*{Code availability statement}
Data are available on \url{https://github.com/pengsl-lab/DEDUCE}.

\section*{Declaration of competing interest}
There are no funds and conflict of interest available for this manuscript.

\section*{Acknowledgment}
\begin{sloppypar}
This work was supported by NSFC-FDCT Grants 62361166662; National Key R\&D Program of China 2023YFC3503400, 2022YFC3400400; Key R\&D Program of Hunan Province 2023GK2004, 2023SK2059, 2023SK2060; Top 10 Technical Key Project in Hunan Province 2023GK1010; Key Technologies R\&D Program of Guangdong Province (2023B1111030004 to FFH). The Funds of State Key Laboratory of Chemo/Biosensing and Chemometrics, the National Supercomputing Center in Changsha (\url{http://nscc.hnu.edu.cn/}), and Peng Cheng Lab.

\end{sloppypar}


\bibliographystyle{unsrt}

\bibliography{REFERENCES1}

\end{document}